\def\year{2025}\relax
\documentclass[letterpaper]{article} 
\usepackage{aaai22}  
\usepackage{natbib}  
\usepackage{times}  
\usepackage{helvet}  
\usepackage{courier}  
\usepackage[hyphens]{url}  
\usepackage{graphicx} 
\usepackage{caption} 
\DeclareCaptionStyle{ruled}{labelfont=normalfont,labelsep=colon,strut=off} 
\usepackage{algorithm}
\usepackage{algorithmic}
\usepackage{microtype}

\usepackage{makecell}
\newcommand{\tabitem}{-- }

\usepackage{tabularx}
\usepackage{makecell}
\usepackage{multirow}
\usepackage{booktabs}
\usepackage{graphicx} 
\usepackage{circuitikz} 
\usetikzlibrary{positioning}
\usepackage{subcaption}
\usepackage{newfloat}
\usepackage{listings}
\definecolor{codegreen}{rgb}{0,0.6,0}
\definecolor{codegray}{rgb}{0.5,0.5,0.5}
\definecolor{codepurple}{rgb}{0.58,0,0.82}
\definecolor{backcolour}{rgb}{0.95,0.95,0.92}
\floatstyle{ruled}
\newfloat{listing}{tb}{lst}{}
\floatname{listing}{Listing}
\title{What Makes a Meme a Meme? \\ Identifying Memes for Memetics-Aware Dataset Creation}
\author {
    Muzhaffar Hazman\textsuperscript{\rm 1}, Susan McKeever\textsuperscript{\rm 2}, Josephine Griffith\textsuperscript{\rm 2}
}
\affiliations{
    \textsuperscript{\rm 1}University of Galway, Galway Ireland.\\
    \textsuperscript{\rm 2}TU Dublin, Dublin Ireland.\\
    m.hazman1@universityofgalway.ie,  susan.mckeever@tudublin.ie, josephine.griffith@universityofgalway.ie
}

\begin{document}

\maketitle

\begin{abstract}
\textbf{\textsc{Warning}: This paper contains memes that may be offensive to some readers.}
\par
Multimodal Internet Memes are now a ubiquitous fixture in online discourse. One strand of meme-based research is the classification of memes according to various affects, such as sentiment and hate, supported by manually compiled meme datasets. Understanding the unique characteristics of memes is crucial for meme classification. Unlike other user-generated content, memes spread via memetics, i.e. the process by which memes are imitated and transformed into symbols used to create new memes. In effect, there exists an ever-evolving pool of visual and linguistic symbols that underpin meme culture and are crucial to interpreting the meaning of individual memes. The current approach of training supervised learning models on static datasets, without taking memetics into account, limits the depth and accuracy of meme interpretation. We argue that meme datasets must contain genuine memes, as defined via memetics, so that effective meme classifiers can be built. In this work, we develop a meme identification protocol which distinguishes meme from non-memetic content by recognising the memetics within it. We apply our protocol to random samplings of the leading 7 meme classification datasets and observe that more than half (50. 4\%) of the evaluated samples were found to contain no signs of memetics. Our work also provides a meme typology grounded in memetics, providing the basis for more effective approaches to the interpretation of memes and the creation of meme datasets.

\end{abstract}

\section{Introduction}



%

Internet Memes have become a staple of expression and interaction within digital communities. Although commonly recognised as multimodal humourous jokes, memes present a medium to participate in a global digital culture \cite{shifman_2014}, spread ideologies \cite{zannettou}, debate political issues \cite{hajizada}, and even distribute propaganda \cite{propmeme}. The expressive nature of memes has led to recent interest in classifying them by the various affects they convey including sentiment, hate, humour, sarcasm and offensiveness. The goals underlying meme classification are varied, e.g.: identifying hateful memes to help kerb the proliferation of hateful rhetoric \cite{hatememesurvey}, detecting signs of cyberbullying \cite{aomd}, recognising trolling behaviour \cite{tamilmeme}, and discerning the sentiment conveyed by memes \cite{memo1_report}.
 
Meme classification datasets provide memes labelled with various affects to aid in the development of meme classifiers via supervised learning. The current state-of-the-art in collecting and labelling sample memes is almost a completely manual process. Since not all content on the Internet are memes, the authors of these datasets have indicated the need to identify memes from non-memetic content during data collection. However, none of the 12 publicly available peer-reviewed meme classification datasets we surveyed disclosed any standards or criteria used to identify memes from non-memetic content. For clarity, we use \textit{meme classification} to describe the task of classifying the various affects conveyed by memes as posed by these datasets, while we refer to \textit{meme identification} as the task of selecting memes from non-memetic content (non-memes). 

Beyond meme classification datasets, work in both the fields of computational and philological research identify, characterise, and interpret memes through the lens of \textit{memetics} \cite{Courtois,shifman_quiddity}, the process by which memes are derived from previous memes through imitation and variation \cite{shifman_2014}. Through memetics, memes form groups with other memes via shared \textit{memetic elements}, which are components or traits common between them. These memetic elements not only indicate that a piece of content has been created from prior memes, but also often encapsulate the meaning and context of a meme \cite{memescape}. The repeated imitation and reuse of memetic elements in millions of memes gives each element its meaning \cite{dubey,zannettou_new}, adding to the constantly evolving collection of visual symbols shared between memes \cite{zannettou}.

\begin{figure}[!ht]
\hspace{0.175\columnwidth}
\subfloat[]{%
  \includegraphics[clip,width=0.3\columnwidth]{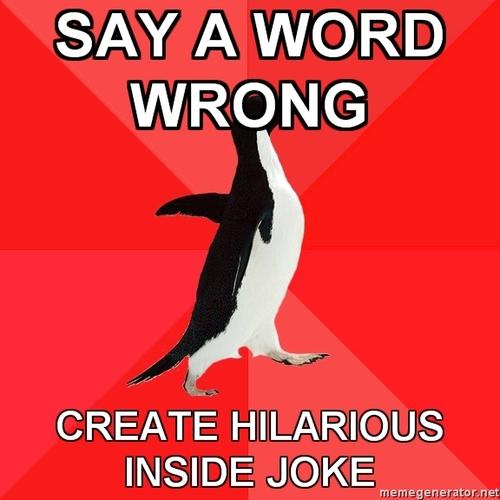}%
}
\hspace{0.05\columnwidth}
\subfloat[]{%
  \includegraphics[clip,width=0.3\columnwidth]{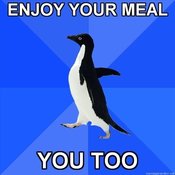}%
}

\caption{Example memes using the popular meme templates: (a) Socially Awesome Penguin and (b) Socially Awkward Penguin}
\label{fig:penguins}

\end{figure}

Interpreting memetic elements is an additional challenge necessary in classifying memes that does not exist when classifying other multimodal content. Understanding the meaning conveyed by these elements is often subtle, and requires familiarity with how each element has been used in preceding memes \cite{memescape}. Consider, for example, the memes in Fig. \ref{fig:penguins}, each using a distinct but similar \textit{meme template} (which \citet{nissenbaum} describes as a recurring image with a text overlay that typically addresses a particular subject or circumstance). The \textit{Socially Awesome Penguin} meme template, as shown in  Fig. \ref{fig:penguins}(a) subverts an awkward social scenario with a charismatic response. This penguin and red background has come to symbolise ``social awesomeness'' and is an inversion of the \textit{Socially Awkward Penguin} meme template\footnote{https://knowyourmeme.com/memes/socially-awkward-penguin}, see Fig. \ref{fig:penguins}(b). The two templates are nearly identical; whether the meme conveys awkwardness or charisma depends solely on which direction the penguin is facing and the colour of its background. The relationship between penguins and either awkwardness or charisma in social situations is obscure. Like many memetic elements, the specific meaning conveyed by these penguins arises solely through their reuse and variation across millions of memes. Similarly to how new memes are derived from previous memes, memetic elements can be derivations of previously established memetic elements, such as how Fig. \ref{fig:penguins}(a) was derived from Fig. \ref{fig:penguins}(b). The new element incorporates some of the meaning of the original element, but is given added meaning through visual editing \cite{zannettou_new}.


In contrast, non-memetic content conveys meaning without relying on memetic elements or on the intended reader being familiar with the larger meme culture. Combining memetic and non-memetic content into meme datasets hinders the development of meme classifiers, as it undermines meme-specific challenges in multimodal affective classification. Current approaches to meme classification are not adequate given that the datasets on which they trained contain both meme and non-meme content. To enable research that addresses challenges unique to meme classification, datasets of genuine verifiable memes are required. To create such datasets, a clear definition of memes and a repeatable methodology for meme identification are needed.

The main contributions of this work are to provide a theoretically grounded definition of memes and repeatable meme identification protocol based on the concept of memetics. We propose that this definition should be used in the creation of meme datasets that are created for the task of affective classification by supervised machine learning models. By surveying the data collection approaches that were used for existing meme classification datasets, we show that none reported any replicable methodology for meme identification or for the filtration of non-memetic content. We present a protocol that identifies a piece of multimodal user-generated content either as a meme or non-meme by verifying its memetic nature. By applying our protocol to samples from the surveyed datasets, we show that current meme classification datasets contain both memes and non-memetic content alike.


Our paper is organised as follows: Section 2 provides an overview of how memes are conceptualised in the literature in two key ways. S2.1 explores how memetic elements signify the memetic nature of every meme. S2.2 illustrates how memetic elements are also used to computationally characterise large collections of memes. Section 3 introduces our novel typology and protocol for meme identification. In Section 4, we present a survey of 12 contemporary meme datasets, revealing that while the authors of these datasets regard memes as being distinct from non-memes, none reported a repeatable protocol for meme identification. Section 5 applies our protocol to seven of these datasets to ascertain whether they solely comprise memes or include non-memetic content. We also analyse instances of non-memes found within meme datasets and discuss the implications of their inclusion in meme datasets. Section 6 positions our work within the context of meme research, outlining its limitations and future research directions.

\section{Background\label{sec:background}}
\subsection{Existing Definitions of Memes}
A meme distinguishes itself from non-memetic content by the way it is created. Unlike other digital content, a meme remains identifiable as a meme and as a member of a larger group of memes due to the observable characteristics it shares with other memes \cite{memescape}. Although each meme is an individual unit of expression, this \textit{shared-ness} exhibits how each meme is derived from other memes, through \textit{memetics} \cite{internet_signs,discursive}.

Memetics in this context entails a large-scale collective process of imitation and variation on popular cultural media. This process is part of a larger participatory culture online \cite{memescape} where popular media is ``\textit{remixed}'' to create a meme that expresses the perspectives of its creator \cite{discursive}. \citet{internet_signs} describes this process as a collective \textit{habit} of \textit{translating} elements from previous memes to create new interpretations, becoming new cultural resources \cite{nissenbaum} that represent shared collective identities \cite{shifman_identity}.

Inspired by several eminent perspectives on memetics, \citet{shifman_2013} reconciled these into the modern digital sense of memes. In doing so, they proposed what has become an influential definition of Internet memes (hereafter, we refer to as ``\textbf{Shifman's definition}''):

\begin{quote}
An internet meme [is defined] as a group of digital items that: (a) share common characteristics of content, form, and/or stance; (b) are created with awareness of each other; and (c) are circulated, imitated, and transformed via the Internet by multiple users.    
\end{quote}


First, note how Shifman spells out the role memetics play in defining Internet memes: every meme is created from ``imitating and transforming'' other memes. The derivation of memes from other memes is central to how memes are created and spread \cite{shifman_2014}. Shifman uses this derivative process to contrast memes against \textit{virals}: singular units of cultural media that, while popular, do not inspire derivatives. Whereas related memes are distinct from one another. Similarly, \citet{memescape} suggest that digital media only become a meme once they generate ``imitations, remixes, and iterations''. This imitation and transformation relates to how  memes are created in groups and that a meme must share ``common characteristics'' with other memes within its group. For example, consider the group of memes in Fig.\ref{img:beskow} which all share the same background. 

Second, the memetics process creates groups of memes that are related by some shared characteristics in one of three ``dimensions'' \cite{shifman_2013}: \textit{Content}, \textit{Form}, and \textit{Stance}. We refer to these components, which are shared and reused between related but distinct memes, as \textit{memetic elements}, such as the penguins in Fig. \ref{fig:penguins} and the reused background in Fig. \ref{img:beskow}. The presence of memetic elements within a piece of digital content indicates that the content was created memetically from a preceding meme, and thus can be verifiably identified as a meme.

Third, we note that modality, in terms of using or combining visual, textual, or verbal elements, does not, in itself, define memes. Although memes are often multimodal, a meme can also be, in its entirety, textual \cite{text_only} or visual \cite{shifman_photos}. Thus, memetics create memes that could manifest as unimodal or multimodal content; conversely, not all multimodal content is memetic.

\begin{table*}[ !ht]
\small
\centering

\begin{tabularx}{\textwidth}{|>{\centering\arraybackslash}X| >{\centering\arraybackslash}X| >{\centering\arraybackslash}X|>{\centering\arraybackslash}X|>{\centering\arraybackslash}X |>{\centering\arraybackslash}X|}
 
 \hline
 Meme Type & \textbf{(a)} Character Macro & \textbf{(b)} Format Macro & \textbf{(c)} Memetic Images & \textbf{(d)} Transferred Symbols & \textbf{(e)} Memetic Trend \\
 \hline

  &  \includegraphics[width = \hsize]{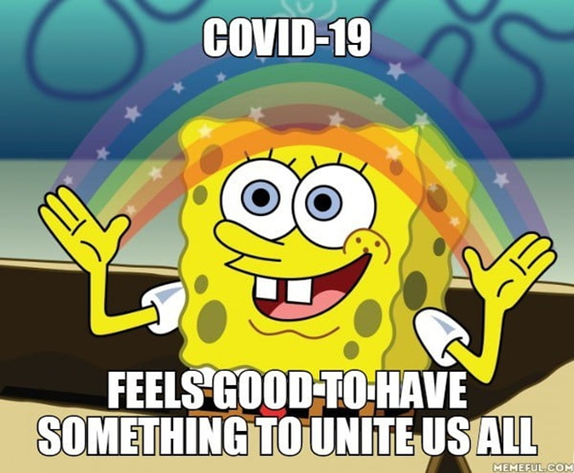} &  \includegraphics[width = \hsize]{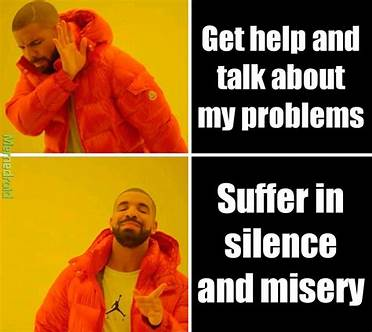} &  \includegraphics[width = 0.9\hsize]{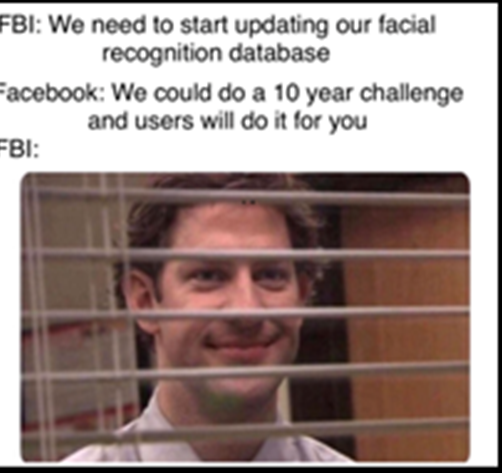} & \includegraphics[width = 0.9\hsize]{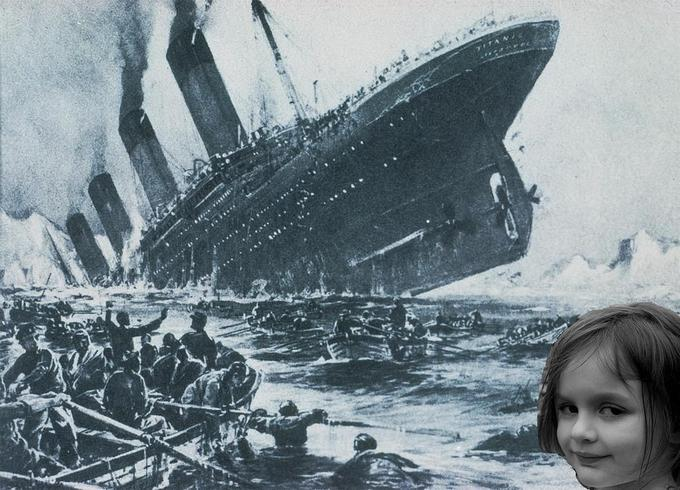} & \includegraphics[width =0.9\hsize]{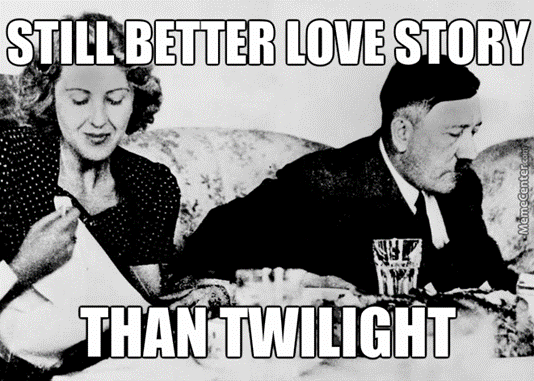}  \\
\hline
\makecell{Memetic Element \\ (Location)} & \textit{Imagination Spongebob} (Background) & \textit{Drake} Template (Background)  &  \textit{Jim Halpert Smiling} (Image) & \textit{Disaster Girl} (Superimposed) & \textit{Better love story than Twilight} (Text)\\
\hline
\makecell{Novel\\ Element} & Text caption & Text clusters & Text caption & Background image& Background image \\
\hline

\end{tabularx}

\caption{Example of each meme type from our meme typology. Each labelled with the \textit{memetic element} (and its location) each shares with, and the \textit{novel elements} that distinguish it from, its \textit{related but distinct} memes (not shown here). \label{tab:types}}

\end{table*}

\subsection{Memetic Elements in Previous Works}
In this section, we highlight several approaches leveraging various types of memetic elements to computationally infer the relationship between one memes and another. Most works that seek to analyse multimodal Internet memes at scale have chosen to focus on features along the \textit{Form} dimension of Shifman's definition to identify shared common characteristics to link related memes. Memetics through this dimension are grounded in a shared visual format or linguistic syntax, i.e. the memes share observable visual, audible, or textual features.

To assess the evolution and influence of memes across various online communities, \citet{zannettou} presents an early example of using shared observable characteristics to link related memes. First, they represented memes based on overall visual similarity. They created vector representations of memes collected from various online communities. They applied a distance metric based on this representation to cluster memes based on visual similarity to a separate set of annotated memes they had collected from meme annotation sites, such as KnowYourMeme.com. Their work also shows how visually similar memes originate from certain online communities and spread through other communities. 

\begin{figure}
\includegraphics[width = \columnwidth]{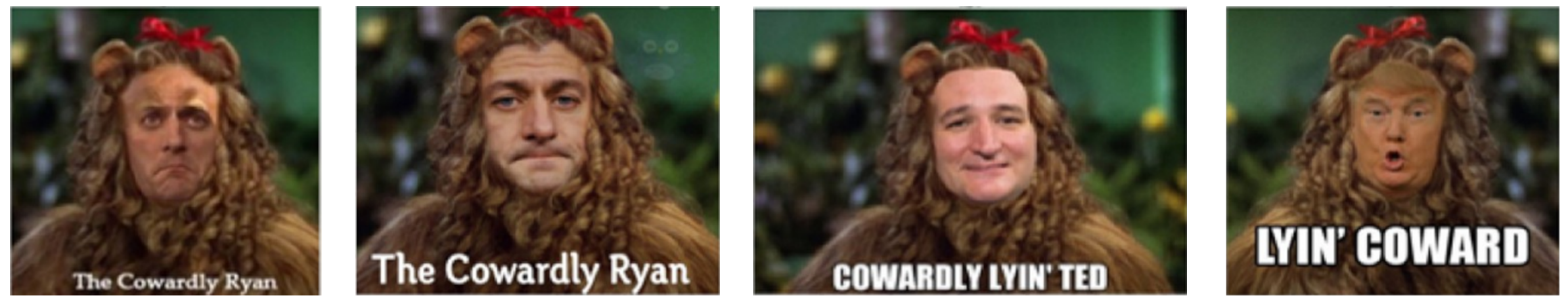}

\caption{Example group of related memes as captured by \citet{politicalmemes}. \label{img:beskow}}
\end{figure}

To track the evolution of memes across political conversations, \citet{politicalmemes} expanded on characterising memes based on overall similarity by incorporating multiple modalities into a deep learning-based representation. They initially trained a classifier that extracts textual and facial features along with visual features to classify memes from non-memetic multimodal content. The authors went on to use the meme representations trained in their classifier to cluster memes based on similar overall appearances. Using this approach, the authors were able to track the evolution of memes in political conversations. While their classifier may appear to solve the task of identifying memes from non-memes, the definition applied by the authors is based solely on the appearance of the text overlays and does not consider memetics. They defined a meme as an image that is either superimposed by text in the Impact font or has text placed in some whitespace over it. Although many memes are consistent with these criteria, this definition strictly limits memes based on the visual appearance of its textual content and does not consider the memetics of individual elements. In fact, within their own illustrated examples, \citet{politicalmemes} published several images which they described as memes, clearly showing memetic imitation via the visual modality, which do not fit either stated criteria (see Fig. \ref{img:beskow}).

\citet{dubey} employed sparse matching of meme templates to establish links between memes. To isolate the background of every meme, they removed overlays and superimposed elements from every meme and compared it to a set of popular meme templates. Clustering memes by their backgrounds allowed the authors to identify memes that shared templates, the memetic elements considered by this approach. The authors were also able to identify memes that used templates that had been remixed with visual alterations and overlays. Although template-based memes are often considered the most common form of memes \cite{knobel}, such a template-based approach limits the possible shared memetic elements to the background of a meme, ignoring how memetic transfer can be observed in foreground elements.

\citet{Courtois} proposed grouping memes based on local visual features as opposed to the overall appearance. This approach recognises visual memetic elements that may be in the template background of one meme but is used as a foreground element in another meme (see example in Table \ref{tab:types}(d) where the character from the \textit{Disaster Girl} meme template\footnote{https://knowyourmeme.com/memes/disaster-girl} is used as a foreground element). They established the relationship between two memes by counting the number of matching features and the difference in position of each matched feature. Much like other works discussed here, this method groups memes by memetic elements. Additionally, this approach acknowledges that visual elements can be replicated as whole-image templates or as individual elements superimposed on another background image, similar to examples discussed in \citet{shifman_photos}.

From these works, we can note two considerations when applying the \textit{ Form} dimension of Shifman's definition at scale: First, the relatedness of a meme to another can be determined by observing shared visual elements between them. Thus, the memetics of a meme can be established by identifying the memetic elements within it that are shared with other memes. Second, these works illustrate three common types of visual memetic elements: \textit{meme templates} as used by \citet{dubey} in which the background is memetically transferred between memes, \textit{foreground visual features} transferred from other memes \cite{Courtois}, and \textit{memetic images} appended with text placed in a whitespace \cite{politicalmemes}.

A limitation of these approaches is that they each rely on a limited pool of memetic elements that was collected from reference memes \cite{dubey,zannettou} or within a collected corpus \cite{Courtois}. Given the dynamic remixing of memetic elements \cite{zannettou_new}, a method is needed to recognise memetic elements based on whether they are used contemporarily in the wider online culture, rather than limited to those from a finite set of reference memes.

\section{Proposed Meme Identification Model}
 Building upon the characteristics of memes as we discussed in the preceding sections, we present the following \textbf{definition for multimodal Internet Memes} which serves as a basis for our meme identification model. 
 \begin{quote}
    A multimodal\footnote{We discuss limitations on modalities in Section 3.3.} user-generated piece of digital content that contains at least one visual or textual component that can be identified as a \textit{memetic element} that is shared with other \textit{related but distinct} memes.
 \end{quote}
 Where we define the following:
 \par \noindent \textbf{\textit{Memetic Element}}: A textual or visual element that is observably reused in multiple distinct memes; e.g. the background shared between multiple memes in Fig. \ref{fig:search}(a). 
 \par \noindent \textbf{\textit{Related but Distinct}}: Pairs or groups of memes that share common memetic elements, but differ from one another by some \textit{novel element}, such as the distinct text captions in Fig. \ref{fig:search}(a) and (b).

 To identify a piece of user-generated content (a ``\textbf{candidate meme}'') as a meme, one would need to observe at least one memetic element within it. For an element in a candidate meme to be considered memetic, we must search for and find \textit{related but distinct} memes using that same element. To recognise such elements, we present a typology of multimodal memes based on the type of memetic element it holds and a protocol to systematically identify such elements.

\subsection{Meme Typology}

\begin{figure}[!t]
\centering
\resizebox{0.7\columnwidth}{!}{
\begin{circuitikz}
\tikzstyle{every node}=[font=\Large ]
\draw [line width=1pt,rounded corners=3.0] (0,12) rectangle  node {\normalsize Multimodal Internet Memes} (4.5,11.25);
\draw [line width=1pt,rounded corners=3.0] (2,10.75) rectangle  node {\normalsize Template Memes} (5.5,10);
\draw [line width=1pt,rounded corners=3.0] (4.25,8.75) rectangle  node {\normalsize Format Macro (FM)} (8,8);
\draw [line width=1pt,rounded corners=3.0] (4.25,9.75) rectangle  node {\normalsize Character Macro (CM)} (8,9);
\draw [line width=1pt,rounded corners=3.0] (2,7.75) rectangle  node {\normalsize Memetic Images (MI)} (6,7);
\draw [line width=1pt,rounded corners=3.0] (2.,6.5) rectangle  node {\normalsize Transferred Symbols (TS)} (6,5.75);
\draw [line width=1pt,rounded corners=3.0] (2.,5.25) rectangle  node {\normalsize Memetic Trend (MT)} (6,4.5);
\draw [->, >=Stealth] (1.25,10.375) .. controls (1.75,10.375) and (1.75,10.375) .. (2,10.375);
\draw [->, >=Stealth] (3.5,9.375) .. controls (4,9.375) and (4,9.375) .. (4.25,9.375);
\draw [->, >=Stealth] (3.5,8.375) .. controls (4,8.375) and (4,8.375) .. (4.25,8.375);
\draw [->, >=Stealth] (1.25,7.375) .. controls (1.75,7.375) and (1.75,7.375) .. (2,7.375);
\draw [->, >=Stealth] (1.25,6.125) .. controls (1.75,6.125) and (1.75,6.125) .. (2,6.125);
\draw [->, >=Stealth] (1.25,4.875) .. controls (1.75,4.875) and (1.75,4.875) .. (2,4.875);
\draw [short] (1.25,11.25) .. controls (1.25,8.25) and (1.25,8.25) .. (1.25,4.875);
\draw [short] (3.5,10) .. controls (3.5,9.25) and (3.5,9.25) .. (3.5,8.375);
\end{circuitikz}
}
\caption{Proposed typology of multimodal Internet memes.}
\label{fig:my_label}
\end{figure}
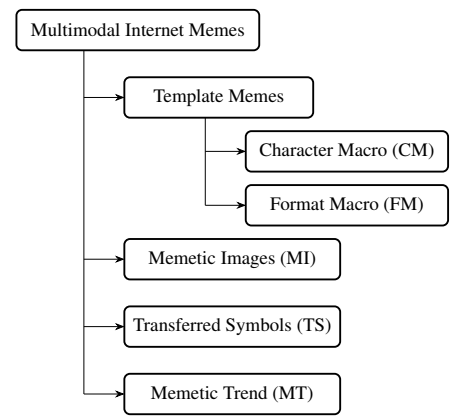

To assist with identifying memetic elements within candidate memes, we present a meme typology based on which component is found to be memetic. 
By combining multiple memetic elements, a meme can fit multiple types simultaneously.
In the context of verifying whether a sample is a meme, identifying memetics in at least one of these categories would establish that it is a meme. Our typology consists of:

\par \noindent \textbf{\textit{Template Memes}} encompass memes that use identical visual backgrounds as templates. These backgrounds are reused across related but distinct memes, often with only the textual content being different. Template Memes are commonly found on \textit{meme generation sites}\footnote{e.g.: \url{https://imgflip.com/}}. Among Template Memes, two common categories exist: \textit{Character Macro} and \textit{Format Macro}, see (a) and (b) in Table \ref{tab:types}, respectively. 
\par \noindent \textbf{\textit{Character Macro} (CM)} memes are established visual templates centred around characters that uses superimposed text captions to signify some ``stereotypical behaviour'' \cite{shifman_photos}. Lists of such characters can be found on meme encyclopaedia sites\footnote{e.g.: \url{https://knowyourmeme.com/}}. 
\par \noindent \textbf{\textit{Format Macro} (FM)} memes use visual templates where the meaning is conveyed through the relative position of visual and textual elements, as described by \citet{hazman_ranlp}. The visual template and the intentional positioning of the superimposed text create metaphors \cite{aomd}, which are then reused by editing the textual content (e.g. Table \ref{tab:types}(b)).
\par \noindent \textbf{\textit{Memetic Images} (MI)} describe memes in which an image is appended with a whitespace containing a text caption as described by \citet{politicalmemes}, e.g.: Table \ref{tab:types}(c). The memetics of such images are realised by applying multiple different text captions to the same image. Similar to template memes, MI memes are visually very similar to their related but distinct memes e.g.: Fig. \ref{fig:search}(c).
\par \noindent \textbf{\textit{Transferred Symbols} (TS)} memes rely on conventionalised visual elements but do not make use of memetic backgrounds. Instead, they use symbols that have gained meaning in the larger meme culture; transferring both the symbol and its adopted meaning onto a new and otherwise non-memetic background. These transferred symbols are usually observed as either a superimposed visual element e.g. in Table \ref{tab:types}(d) or as a visual segment e.g. Fig \ref{fig:search}(b). \par \noindent \textbf{\textit{Memetic Trend} (MT)} memes are perhaps the most difficult to identify. These memes usually share similar text but may not use visually memetic elements. These are created in response to a participative trend or topic that inspired large amounts of visually dissimilar memes, such as the \textit{Better Love Story than Twilight} memetic trend shown in Table \ref{tab:types}(e).

\subsection{Meme Identification Protocol}
 We present our protocol for multimodal meme identification below. First, we define \textbf{$C$} as a candidate multimodal meme: an image file that contains visual and textual elements that we would like to identify as either a meme or as non-memetic. Further, we define \textbf{$IS(image)$} as a \textit{reverse image search} where image file, $image$, is submitted as a query to the search engine, which returns $N$ visually similar image files. \textbf{$TS(text)$} represents a traditional image search function that returns $N$ image files that correspond to, or contains, a  given text input, $text$. When reviewing the results of these search functions, the objective is to find at least one meme that is \textit{Related to but Distinct} from $C$, i.e.: image files that share at least one common memetic element with $C$ while being distinct from each other by some novel element(s). For both $IS$ and $TS$, we reviewed all returned search results, which proved to be manageably few. Our protocol consists of the following steps:

\begin{figure}[ !ht]
\centering
\subfloat[]{%
  \includegraphics[clip,width=0.8\columnwidth]{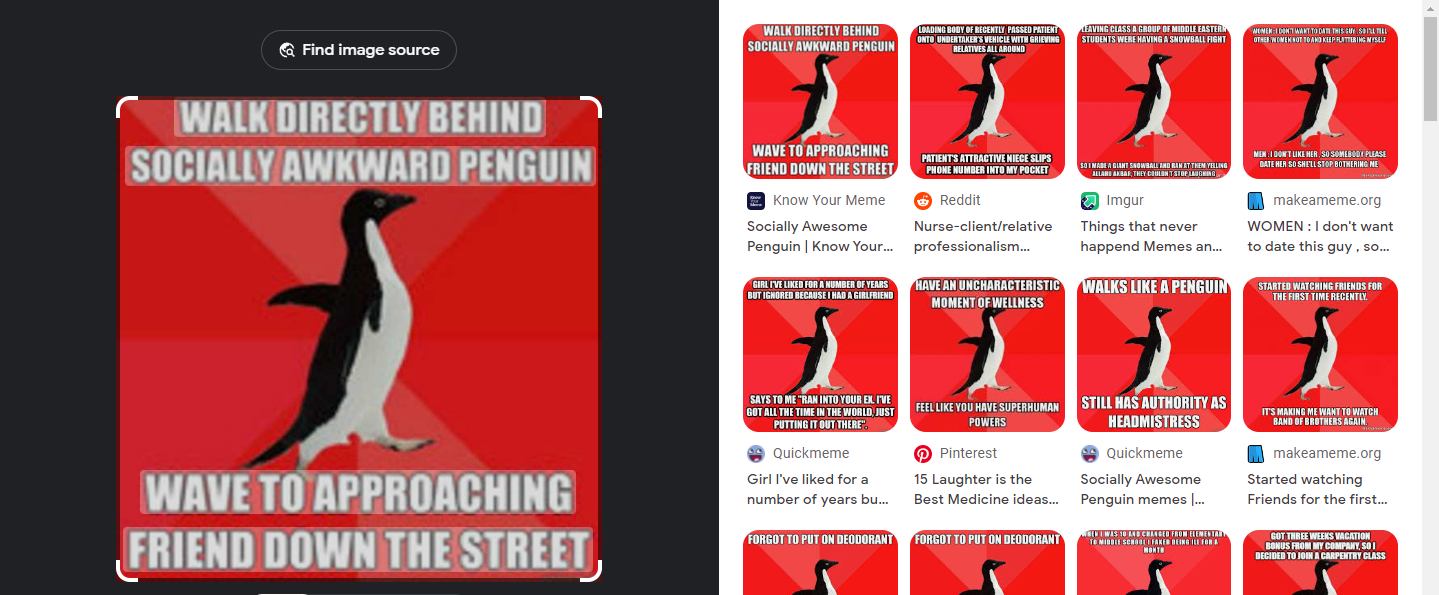}%
}

\subfloat[]{%
  \includegraphics[clip,width=0.8\columnwidth]{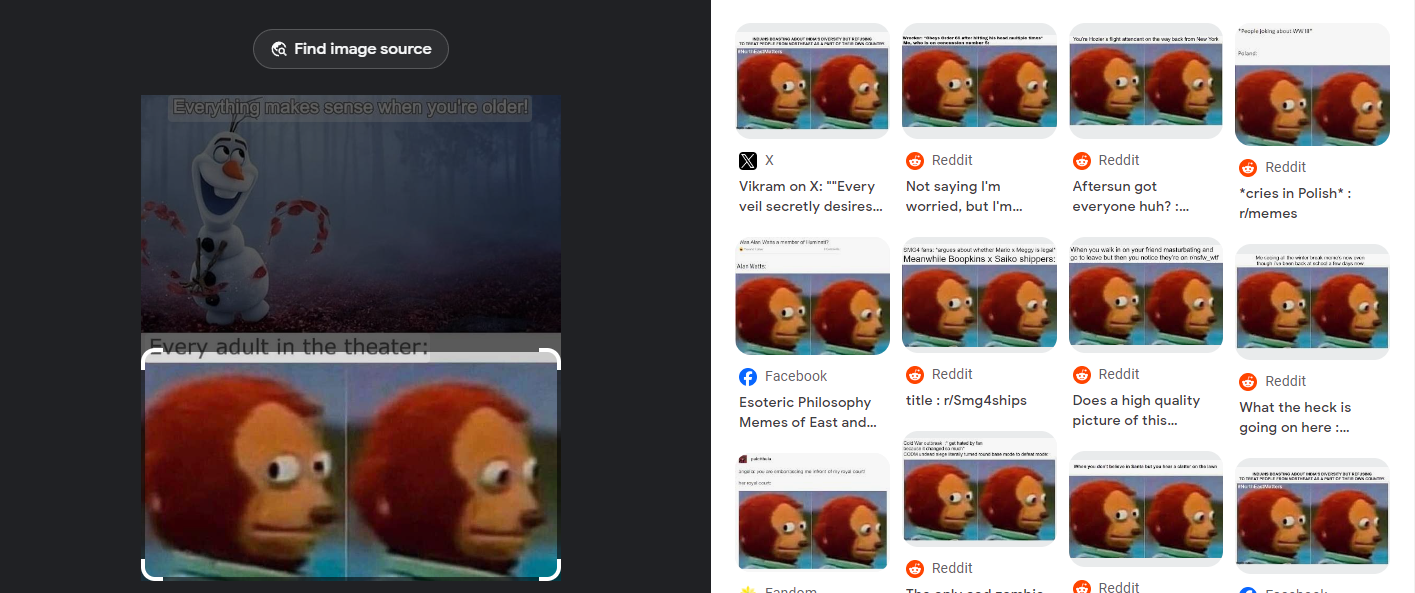}%
}

\subfloat[]{%
  \includegraphics[clip,width=0.8\columnwidth]{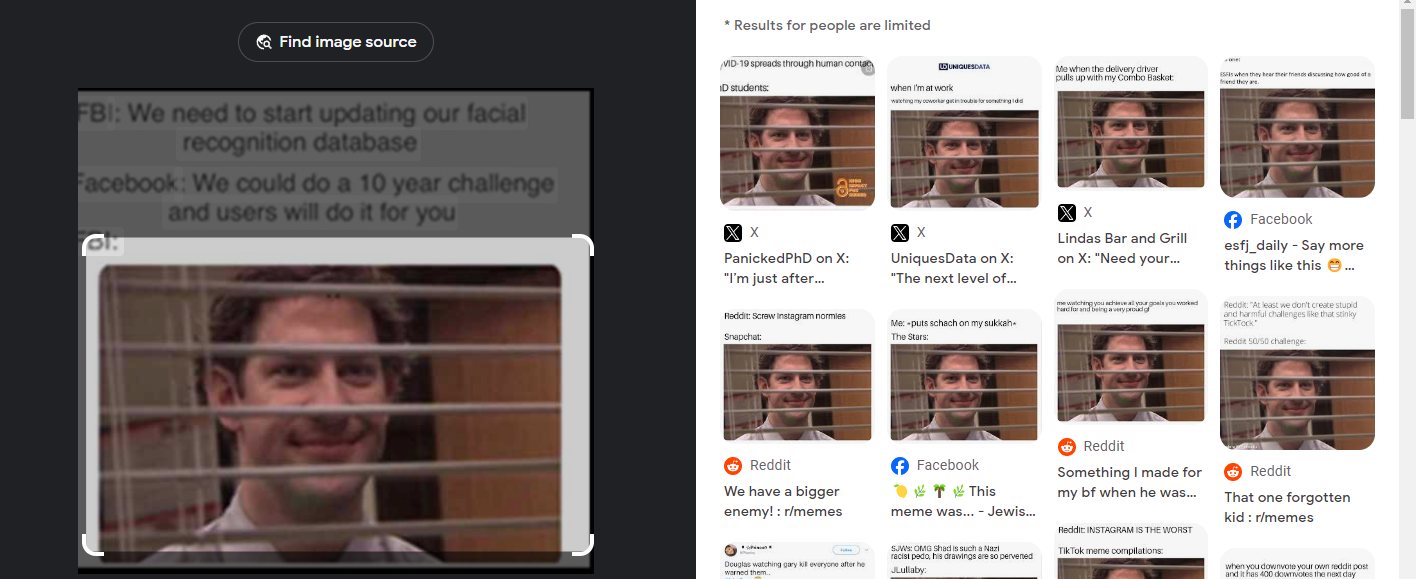}%
}
\caption{Google Image Search\footnote{https://images.google.com/} results for (a) a Character Macro meme showing related memes using the same template, (b) Transferred Symbols meme, where the search input is cropped to the bottom segment, showing related memes using the \textit{Monkey Puppet} memetic element and (c) Memetic Image meme, where the search input is cropped to remove the textual element, showing related memes that reuse the same image.}

\label{fig:search}

\end{figure}

\footnotetext{https://images.google.com/}
\begin{enumerate}
    \item Retrieve images that are visually similar to $C$, $IS(C)$ and review these for memes that share the same background as $C$ but with different foreground text or visual content. If at least one is found, $C$ is a \textit{Template Meme}. Else: go to Step 2.
    \item Consider the overall visual format of $C$. If $C$ comprises a character in the centre with a text caption overlay (e.g. Fig \ref{fig:search}(a)), go to Step 3. If $C$ comprises of multiple segments, each containing a separate visual content (e.g. Fig. \ref{fig:search}(b)), go to Step 4. Otherwise, if $C$ comprises an image and a separately appended whitespace (e.g. Fig. \ref{fig:search}(c)), go to Step 5. Otherwise, go to Step 6.
    \item Crop $C$ to remove textual content, creating $c_{crop}$. Review $IS(c_{crop})$ for any meme that shares the same background as $C$ but with different foreground text or visual content. If found, $C$ is a \textit{Character Macro} Meme. Otherwise, go to Step 6.
    \item Crop $C$ into its visual segments, $c_m$, where $m$ ranges from 1 to the number of segments in $C$. Review $IS(c_m)$ for each $m$, and search for memes that pair segment $m$ with other visual segments or different text captions. If found for any $c_m$, $C$ is a \textit{Transferred Symbols} meme.  If none is found for all $c_m$, go to Step 6.
    \item Crop $C$ to remove the white space to create $c_{img}$.  Review $IS(c_{img})$ for memes that contain $c_{img}$ but have different textual content. If found for any $c_{img}$, $C$ is a \textit{Memetic Image} meme. Otherwise, go to Step 6
    \item Crop each distinct visual elements particularly those that appear to have been superimposed onto a background image, to create $c_{elem}$. Review $IS(c_{elem})$ for other memes that include $c_{elem}$. If found for any $c_{elem}$, $C$ is a \textit{Transferred Symbols} meme. Otherwise, go to Step 7.
    \item Extract the textual elements from within $C$, $c_{text}$. Review $TS(c_{text})$ for distinct memes that contain textual content that is, wholly or in most parts, identical to that in the candidate meme. If $c_{text}$ is shared across numerous
    visually dissimilar memes,  $C$ is a \textit{Memetic Trend} meme. Otherwise, go to Step 8.
    \item If the candidate meme does not contain any identifiable memetic elements, as per the steps above, it is considered non-memetic.
    
\end{enumerate}




\begin{table*}[!ht]
\centering
    \begin{tabular}{|>{\raggedright\arraybackslash}p{0.2\textwidth} | p{0.35\textwidth} | p{0.35\textwidth} | }
    \hline
     \textbf{Dataset} &\textbf{Meme Identification Reported?:\newline
     Approach}  & \textbf{Reported Filtration Criteria}  \normalsize
 \\
 
        \hline

         \citet{Miliani2020DANKMEMESE}&
 
         Yes: Annotated by labelers based on: ``formal aspects (layout, multimodality and manipulation) as well as content, e.g. ironic intent''.
                  &
         \tabitem Relevance to selected topic.\newline
         \tabitem Italian-only text. \newline
         \\
         \hline
        
         \citet{memo1_report} &
          Yes: Manual selection by authors.   &
         \tabitem Contains ``clear background picture'', and ``embedded textual content''. \newline
         \tabitem English-only text \\
 \hline
        \citet{memo2_report}&
         Yes: Manual selection by authors and ``extensive cleaning'' of scraped content by authors.
         &
     \tabitem  Collected from ``public domains''

         \\
         \hline
         \citet{hateful_data}&  
         Yes: Manual selection by authors.
         &  
         \tabitem The original image in the meme must be replacable while preserving the meme's meaning.
         \\
         \hline
         \citet{harmeme} &
        Yes: Keyword search appended with ``memes'' followed by manual selection by authors.
        &
         \tabitem English-only text. \newline
\tabitem Text is readable. \newline
\tabitem  Contains both text and visual content. \newline
\tabitem Does not contain \textit{cartoon} visuals.
         \\
         \hline
         \citet{propmeme}&
                 Yes: Manual selection by authors. &
         \tabitem Does not contain ``diagrams/graphs/tables'' or \textit{cartoon} visuals \newline
\tabitem  Contains both text and visual content.
         \\
         \hline
         \citet{tamilmeme}& 
         Yes: Volunteers were  requested to submit memes specifically. & \tabitem Includes Tamil-only text.
         \\

         \hline             

         \citet{totaldefmemes} &Yes: Keyword search appended with ``memes'', followed by filtration by annotators.  & 
         \tabitem Minimum image resolution. \newline
         \tabitem Maximum text length. \newline
         \tabitem Contains text.

         \\
         \hline

          \citet{fersini}&Yes: Manual selection by authors including from meme creation sites.\newline
         & 
        \tabitem N/A
         \\
         \hline
         \citet{memo3_data}&
         Yes: Manual selection by authors.& 
        \tabitem Text is in code-mixed Hindi-English.         
         \\
         \hline
         \citet{memosen}&
         Yes: Content search appended with keyword ``memes'' followed by manual selection by authors. & 
        \tabitem Bengali-only text. \newline
        \tabitem Text is readable. \newline
\tabitem  Contains both text and visual content. \newline
\tabitem Does not contain \textit{cartoon} visuals.
         \\
         \hline

      \citet{mute} &
         Yes: Manual selection by authors from content search appended with keyword ``memes'' and ``popular meme pages''.\newline & 
        \tabitem Text is in code-mixed Bengali-English.\newline
        \tabitem Contains both text and visual content.  \newline
        \tabitem Text and visual content are clear. \newline
        \tabitem Does not contain \textit{cartoon} visuals.
        \\
    \hline
    \end{tabular}
    \caption{Meme Identification Approaches and Reported Filtration Criteria used by current meme classification datasets. For the classification task and the size of each dataset, see Appendix.}
    
    \label{tab:datasets}
\end{table*}

\subsection{Addressing Current Challenges in Meme Identification}
Basing meme identification on the existence of \textit{related but distinct} memes, as outlined in our protocol, removes the reliance on personal judgement or familiarity with meme culture from the process of recognising memetic elements as was used in some recent datasets \cite{Miliani2020DANKMEMESE,tamilmeme}. Due to the extremely large number of visual and textual elements that may possibly be memetic, no annotator could be expected to recognise all such elements from personal knowledge alone. Other alternatives to meme identification have attempted to leverage reference memes sourced from community-maintained encyclopaedias of popular memes \cite{IMKG} or randomly sampled from social media platforms \cite{sheratt_multi}. However, since memetic elements are constantly added to meme parlance \cite{memescape}, it is likely impossible to prepare a set of reference memes that is large enough to encompass the entirety of meme culture. 

Arbitrary collections of reference memes, whether formally collected or within the personal familiarity of an annotator, impose artificial constraints on the potential diversity of memetic elements. Both approaches rely on memetic elements achieving some degree of global popularity and are likely to overlook two significant categories of memes: non-English and newly emerging memes. Due to being more culturally specific, memes that are rooted in non-English cultures or use non-English text are less likely to achieve the global popularity required to be included into a meme encyclopaedia or to be represented within a random sampling of a meme-sharing communities which tend to use English as their lingua franca. Furthermore, since a meme only requires one memetic derivative to qualify as a meme, these approaches are likely to overlook newly emerging memes that have yet to achieve significant popularity.

To allow us to recognise personally unfamiliar, culturally specific, as well as emerging memetic elements, our protocol instead employs visual search engines. Specifically, we use Google Image Search to retrieve visually similar images and/or multimodal memes, which are then assessed for memetic elements shared with the candidate meme. If any of the results share memetic element(s) with the candidate meme but is otherwise distinct, it qualifies as a \textit{related but distinct} meme. Candidate memes for which the search only returns results identical to itself, even from multiple different sources, indicate that the candidate meme is likely a viral piece, rather than a memetic one, as described by \citet{shifman_2014}. 

Our protocol caters specifically for multimodal memes as they are represented within contemporary multimodal meme datasets: where each sample comprises an image file that contains both visual and textual content. We acknowledge that this representation does not encompass all forms of memes or multimodality within memes. For example, a \textit{Memetic Image} meme could manifest as a memetically reused image accompanied by novel text captions within the body of a social media post; instead of textual content within the image file. Despite differences in format, it combines the novel and memetic elements in the same way that a Memetic Image meme (e.g. Fig. 5(c)) does and is both multimodal and memetic. However, since current meme datasets do not capture textual content outside of image files, neither does our protocol. 



\section{Analysis of Current Meme Datasets}
To assess the current practice of meme identification, we examine how current meme datasets handle the task of meme identification from two perspectives. Each dataset reportedly consists exclusively of memes, each having been manually selected and annotated with affective classification label(s). First, we present a survey of current approaches to meme identification taken by the authors of current meme classification datasets. Here, we attempt to identify (1) whether current datasets regard memes as distinct from non-memes and (2) the criteria and methods used to identify the memes selected for each dataset. Specifically, we examine publications that present memes as samples in a dataset, each annotated with one or more affective classification label(s). We excluded publications presenting datasets that are re-annotations or extensions of another dataset or those that lack scientific peer review. Second, we apply our protocol to samples taken from the datasets. This allows us to determine whether current meme datasets include both memes and non-memes.

\subsection{Meme Identification Approaches in Existing Meme Classification Datasets}
We reviewed the publications that accompany 12 meme datasets. The purpose of examining these datasets is to establish how memes are currently being defined in the field of meme classification. We sought information on any steps taken to either identify memes or to remove non-memetic content during data collection. Then, we extracted any criteria reportedly used during data collection, focussing on whether each criterion helps distinguish memetics within memes. Table \ref{tab:datasets} summarises our findings on: (1) the presence of a meme identification step, and (2) any criteria used to qualify memes for inclusion in each dataset.

\subsubsection{Memes as Identifiable Content}
As shown in Table \ref{tab:datasets}, all 12 dataset publications treat memes as a distinct type of content and have reported some additional effort to filter memes from a large collection of image files. All reviewed works opted for manual approaches to meme identification. Most frequently, the authors of each dataset reported manually selecting memes from image search results, web scraping, and/or browsing online social groups \cite{memo1_report,memo2_report,hateful_data,harmeme,momenta,fersini}. However, none of these works reported criteria for how memes were chosen during this manual selection process. \citet{memo2_report} also reportedly performed an ``extensive cleaning'' step following their data collection via web-crawling but neither the purpose nor the criteria used for this step were reported.

Furthermore, \citet{harmeme,momenta,memosen,totaldefmemes} used search functions to assist in filtering out non-memes by adding the keyword ``meme'' to their image search input. 
This delegation of the meme identification task to search engines is perhaps best demonstrated by \citet{pixels}, who instead of publishing the data used to train their classifiers, uploaded code that  sends ``meme''-appended queries\footnote{Queries used: \textit{racist meme}, \textit{jew meme}, and \textit{muslim meme}.} to Google Image Search and downloads the returned results. 

Three publications delegated meme identification to annotators: \citet{totaldefmemes} asked their annotators if they considered each sample to be a meme. \citet{Miliani2020DANKMEMESE} included \texttt{Meme} versus \texttt{notMeme} as a classification label, where the authors reported that these labels were applied based on ``formal aspects (such as layout, multimodality and manipulation) as well as content (e.g. ironic intent)''\footnote{No further details were provided.}. \citet{tamilmeme} prompted public participants to submit memes. However, none of these three works published any annotator guidelines on what constitutes a meme nor any protocol to recognise indications of memetics.

\begin{table*}[!t]
\centering
\begin{tabular}{|l|llllll|lll|}
\hline
\textbf{Dataset} & \multicolumn{6}{l|}{\textbf{Memes} - by type (\% of total sample)} & \multicolumn{3}{l|}{\textbf{Non-Memes} (\% of total sample)} \\
 &  CM & FM & MI & TS & MT & \textbf{Total} & nMIT &  nMM &  \textbf{Total}\\
 \hline
\citet{memo1_report} & 33 & 7 & 7 &  6  & 2 & \textbf{55} & 39  & 6 & \textbf{45}\\
\citet{memo2_report} & 27  & 19  & 20 & 9 & 2 & \textbf{77} & 21  &  2 & \textbf{23} \\
\citet{hateful_data} & 5 & 0 & 7 & 3 & 1 & \textbf{16} & 80 & 4  & \textbf{84} \\
 \citet{harmeme} & 14  & 6  & 18 & 6 & 3 & \textbf{48} & 48& 4   & \textbf{52} \\
 \citet{propmeme} & 15 & 7 & 9 & 7 & 3 & \textbf{41} & 58 & 2  & \textbf{60}  \\
 \citet{totaldefmemes} & 10  & 16 & 10  &  9 & 4 & \textbf{49} & 33 & 18  & \textbf{51}  \\
\citet{fersini} & 20  & 12 & 18 & 7 & 3 & \textbf{62} & 38 & 0  & \textbf{38}  \\
\multicolumn{1}{|r|}{\textbf{Average}} &   &  &  &  &  & \textbf{49.6} &  &   & \textbf{50.4}  \\
\hline
\end{tabular}
\caption{Proportion of samples (200 per dataset) found to contain memetic elements by type: CM: Character Macro memes, FM: Format Macro memes, MI: Memetic Images, TS: Transferred Symbols memes, MT: Memetic trend memes; and proportion of samples not containing any memetic elements: nMIT: Non-Memetic Image-Text multimodal content, nMM: non-memetic non-multimodal content. For each dataset's classification task, size and distribution licence, see Appendix.\label{tab:results}}
\end{table*}

\subsubsection{Filtration Criteria}
The rightmost column in Table \ref{tab:datasets} shows the filtration criteria that were applied during the creation of each dataset. Although each criterion characterises the samples included in each meme dataset, none of the criteria directly relate to memetics nor do they appear to outline what constitutes a meme. In lieu of memetics, we expected to identify the authors of each dataset to determine whether a sample constitutes a meme. Instead, none of the criteria reported reveal how the authors of these datasets identified memes from non-memes. Furthermore, the criteria reported varied significantly between the datasets, and do not collectively point towards a common set of characteristics used to identify memes. 

Most frequently, dataset authors restricted their collection according to multimodality by excluding samples that included only image or text content \cite{Miliani2020DANKMEMESE,memo1_report,memo2_report,harmeme,momenta,mute,memosen}. Across all these datasets, multimodality is represented as text that appears alongside visual elements within a single image file. Although most of these works assert that multimodality is a common characteristic of Internet memes (and restricted their datasets to multimodal samples out of consistency), \citet{propmeme,momenta} go further by incorporating multimodality into their definitions of memes; both defining memes specifically as an image with appended/superimposed text. As we have noted previously, multimodality alone, without considering memetics, does not make a meme \cite{shifman_2013}. While other works analyse unimodal text-only memes \cite{text_only} and image-only memes \cite{shifman_photos}, our analysis is entirely focused on multimodal meme data sets.

Furthermore, several data sets also restricted their data collection based on language \cite{Miliani2020DANKMEMESE,memo1_report,hateful_data,harmeme,memosen,mute,memo3_data}. However, this did not appear to be related to how the authors identified memes from non-memetic content. Some chose to exclude samples that contained ``cartoon'' visual elements \cite{harmeme,propmeme,memosen,mute}. This decision seems to originate from the assumption that machine learning models cannot effectively handle cartoon visual elements \cite{harmeme}.

\subsubsection{Summary} The authors of current meme datasets refer to memes as a distinct and identifiable form of digital content, incorporating additional effort to filter memes from non-memetic content during data collection. However, none reported a replicable and theoretically grounded methodology to do so. Although these works report various filtration criteria used during data collection, none seem to aid in distinguishing memes from non-memetic content. In fact, these works do not appear to consider memetics during meme collection.  Furthermore, none of the works discussed the role of memetics during the labelling process or how annotators were asked to handle the interpretation of unfamiliar memetic elements. Current approaches rely heavily on manual labour and subjective personal judgments. Therefore, the current approaches and criteria used to create current meme datasets do not offer any alternatives to our meme identification model and identification protocol. As long as data collection methods for creating meme datasets do not apply a clear standard for meme identification, the performance of models built on such datasets may reflect the state of the art in multimodal classification, but not necessarily that of meme classification.

\subsection{Identifying Non-Memes in Current Meme Datasets using Our Protocol}
 This section assesses whether the lack of criteria for identifying memes (as highlighted in the previous section) leads to the existence of non-memes in current datasets. To do this, we manually apply our protocol to samples from each of the meme datasets that include only English text. Since our protocol requires recognising \textit{Memetic Trend} memes via textual elements of memes, we excluded datasets that were specifically filtered for memes with text in languages other than English \cite{Miliani2020DANKMEMESE}, \cite{tamilmeme}, \cite{memosen}, \cite{mute}, \cite{memo3_data}.
 
 Since we performed this evaluation manually, it was only viable for us to take a small sample from each dataset. We note that for some of the datasets, this sample size represents a smaller portion than for others (see Appendix for a list each dataset size). We randomly sampled 200 memes from each dataset using a Python script (see Appendix). For datasets that included splits or multiple subsets, each subset was sampled with the same probability. Our sampling did not take into account any classification labels or contents of the image files as all the samples are equally asserted to be memes by their respective authors regardless of label. As current datasets report to already exclude non-memes, one would expect to only find memes in such a small sampling. Thus, we argue that a substantial presence of non-memes in our sample is indicative of a wider presence of non-memes in the full dataset.

Applying our protocol to each sample chosen, we identify it as one of the following:
\begin{itemize}
    \item non-memetic image-text multimodal content (as \textbf{nMIT} in Table \ref{tab:results});
    \item non-multimodal text-only or visual-only (as \textbf{nMM} in Table \ref{tab:results}); or
    \item for samples that are identified as memes, we record its type per our meme typology.
\end{itemize}

\subsubsection{Results}
Table \ref{tab:results} presents the proportions of samples from each dataset identified either as a meme or as a non-meme (as nMIT or nMM). On average, 50.4\% of the content sampled from each dataset was found not to contain identifiable memetic elements. This proportion varies widely between the different datasets, with a minimum non-meme proportion of 23\%. These findings show that all the evaluated datasets contain a sizable number of samples without memetic elements and therefore are not memes. Combined with the lack of reported criteria used in relation to meme identification, it appears that current meme datasets do not systematically distinguish memes from non-memes. Further, it is unclear whether memetics was considered at all during the creation of these meme datasets, and if so, what definition of memetics was used and applied.

\subsubsection{Discussion}
\begin{figure}[!t]
\centering
\begin{subfigure}{.4\columnwidth}
  \centering
  \includegraphics[width=0.85\textwidth]{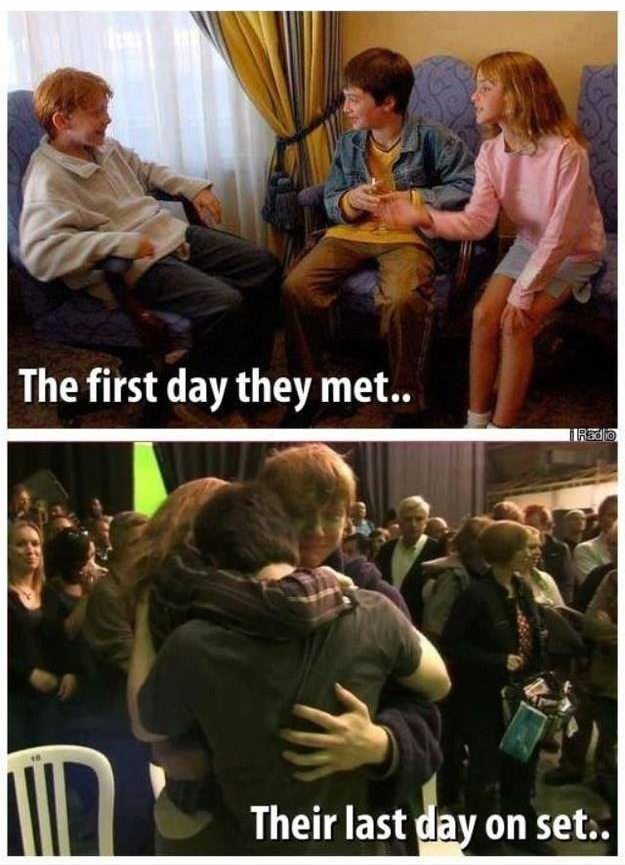}
  \caption{}
\end{subfigure}%
\hfill
\begin{subfigure}{.4\columnwidth}
  \centering
  \includegraphics[width=0.7\textwidth]{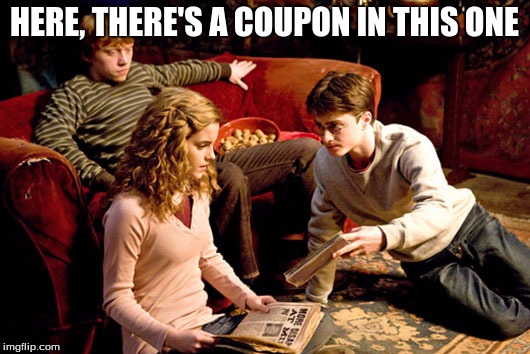}
  \caption{}
  \centering
  \includegraphics[width=0.7\textwidth]{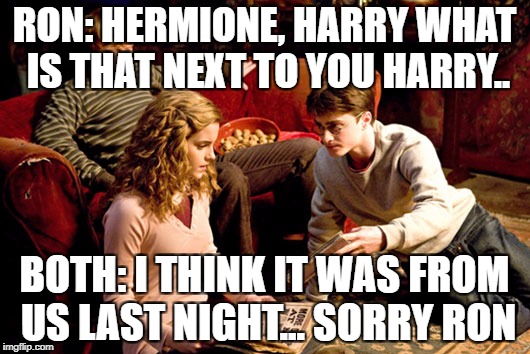}
  \caption{}
  
\end{subfigure}
\caption{(a) Non-meme sampled from \cite{memo1_report}; (b) \& (c) Example of related but distinct meme pair that share common albeit unpopular visual background.}
\label{fig:harry}
\end{figure}

Consider the sample found in the \cite{memo1_report} dataset: Fig. \ref{fig:harry}(a). Although this sample fulfils all the criteria stated by the authors of that dataset viz.  contains a ``clear background image'' and ``embedded textual content'', which is ``wholly in English'', it does not contain any memetic elements. When evaluated through our protocol, the reverse image search returned numerous identical copies, indicating that it had been widely propagated across the Internet in its original form. As we discussed previously, this fits into Shifman's \citet{shifman_2014} definition for a ``viral'', but does not contain any elements that have been reused memetically prior or since. Thus, understanding the meaning conveyed by this sample requires some cultural context but does not require any interpretation of memetic elements. In contrast, the same three people are included in Fig. \ref{fig:harry}(b), a Template Meme for which we found only two related but distinct memes, including the one shown in Fig. \ref{fig:harry}(c). The existence of these memes is proof that the background is a memetic element. Furthermore, this exemplifies how our protocol can identify memetic elements that have not achieved significant popularity.

\begin{figure*}[!t]
\centering
\begin{subfigure}{.3\textwidth}
  \centering
  \includegraphics[width=0.8\linewidth]{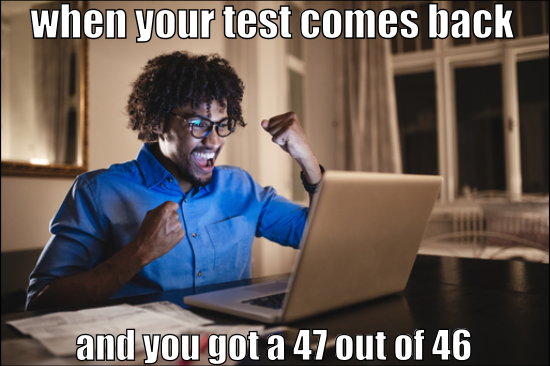}
  \caption{}
  \label{fig:chromosome_synth}
\end{subfigure}%
\begin{subfigure}{0.3\textwidth}
  \centering
  \includegraphics[width=0.4\linewidth]{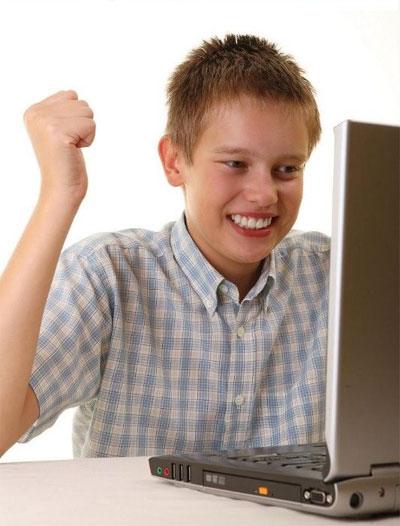}
  \caption{}
  \label{fig:firstday}
\end{subfigure}
\begin{subfigure}{0.3\textwidth}
  \centering
  \includegraphics[width=0.7\linewidth]{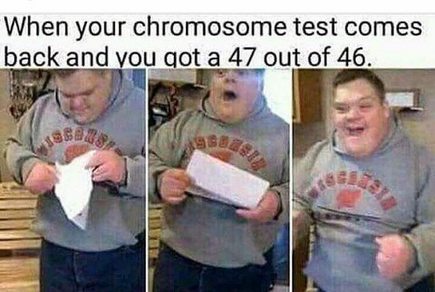}
  \caption{}
  \label{fig:47}
\end{subfigure}
\caption{(a) Synthetically recreated meme (\textcopyright Getty Images) sampled from \cite{hateful_data}; (b) \textit{First Day on the Internet Kid} Character Macro template; and (c) Example meme from the derogatory ``47 out of 46'' Memetic Trend.}
\label{fig:chromosome}
\end{figure*}

Other than the inclusion of memetic elements, we noted that memetic and non-memetic samples appear superficially very similar. Like many multimodal memes, the sample shown in Fig. \ref{fig:harry}(a) utilise visual segments to create comic-like frames, similar to memes shown in Fig. \ref{fig:search}(b) and Table \ref{tab:types}(b). Also, many non-memetic samples use visual formats used in Character Macro and Memetic Images memes, but with images that have not inspired derivative memes. These similarities highlight the challenge of identifying memes from non-memetic content without systematic steps to verify the memetic nature of the contents within a meme, and referring to large collections of memetic elements. 

Further, we also observed that all samples, memetic and otherwise, almost universally include cultural references. As such, cultural contextual information is needed to classify the affect of both types of content, such as that proposed by \citet{momenta}. However, within non-memetic content, these references can be contextualised by recognising what \citet{knowmeme} describes as ``commonsense knowledge'' that underlie the symbolic meanings of elements within multimodal content. Crucially, \textit{commonsense} here describes facts that an average person is expected to know \cite{knowmeme}. In contrast, the memetic process itself serves as the cultural context for memetic elements \cite{internet_signs}, necessitating familiarity with the meme culture to be able to interpret such elements \cite{memescape}. Since memes combine references to both ongoing cultural events and meme culture in particular, interpreting memes requires context and knowledge from beyond \textit{commonsense} alone, requiring knowledge of the memetic elements within the meme \cite{dubey}.

Furthermore, consider the sample shown in Fig. \ref{fig:chromosome}(a), sampled from the \citet{hateful_data} dataset. This example illustrates the nuanced semantics conveyed by memetic elements. Citing concerns over copyright infringement, the authors of this dataset tasked annotators with recreating memes using ``near-identical'' visual content; often removing visual memetic elements in the process. This sample depicts a person excited to receive a result of ``47 out of 46'' for some undefined test. It is ostensibly joyful and is labelled as non-hateful in the dataset. However, we believe that two memetic elements were removed during the recreation process, resulting in a loss of meaning of the original meme. First, we believe that the original meme used the Character Macro template \textit{First Day on the Internet Kid}, see Fig. \ref{fig:firstday}. Unlike the background of the recreated sample, this character has come to symbolise naive excitement for a false, deceitful, or otherwise unexciting occurrence online. This suggests a more subtle (and possibly negative) meaning of the test results mentioned here. Second, applying Step 7 of our proposed protocol reveals that the original meme was likely part of a Memetic Trend surrounding memes with captions including ``47 out of 46''. These numbers refer to chromosomes and, as part of a memetic trend, were often appended derogatorily to images depicting persons with Trisomy or Down Syndrome; see example in Fig. \ref{fig:47}. Without recognising these memetic elements, which were possibly obfuscated by the recreation process, this meme takes on a harmless and joyful appearance. Only when viewed through the relevant memetic contexts can we clearly recognise the derogatory ``humour'' being conveyed.
\par Since current datasets contain both memetic and non-memetic content, they may be well suited for training supervised learning models to classify multimodal content in general, but not specifically memes. The lack of clear delineation in terms of memetics or any reported criteria used to differentiate memes from non-memes  in these datasets begs the question: what qualifies these datasets as ``meme datasets''? If our protocol had been applied during the creation of these datasets, each sample would be systematically and verifiably identified as a meme. This, in turn, would allow the distinction between meme classification and multimodal content classification to be clearly delineated; where meme classification specifically requires the interpretation of memetic elements.

\section{Limitations \& Future Works}
In line with current practices in preparing meme datasets, our protocol is designed to be performed manually. However, this severely limited the number of samples we could evaluate per dataset. Although this allowed us to observe that current datasets do not exclude non-memetic content, the proportions of memes to non-memes we reported in Table \ref{tab:results} may not hold across the entirety of each dataset, and we caution against generalising our observations as such.

Although this work does not directly propose an automated meme identification solution, our protocol illustrates three core capabilities required by any such solutions. Firstly, the solution must recognise memetic elements both globally and locally \cite{Courtois}. Secondly, it should access a vast repository of digital artefacts to recognise memetic elements within emerging and culturally specific memes. To achieve this, Image Search APIs, such as Google Image Search, could be used instead of our manual interaction with Google Image Search. Lastly, an automated solution cannot rely on matching candidate memes to other memes on similarity alone, but must also be able to recognise both novelty between related-but-distinct memes. Additionally, to build machine learning-based solution, a dataset of systematically labelled memes and non-memes is needed. Our protocol provides a repeatable and verifiable approach to collecting such data.

\begin{figure}[!t]
\captionsetup[subfigure]{labelformat=empty}
\begin{subfigure}{.5\columnwidth}
\centering
\includegraphics[height=3cm]{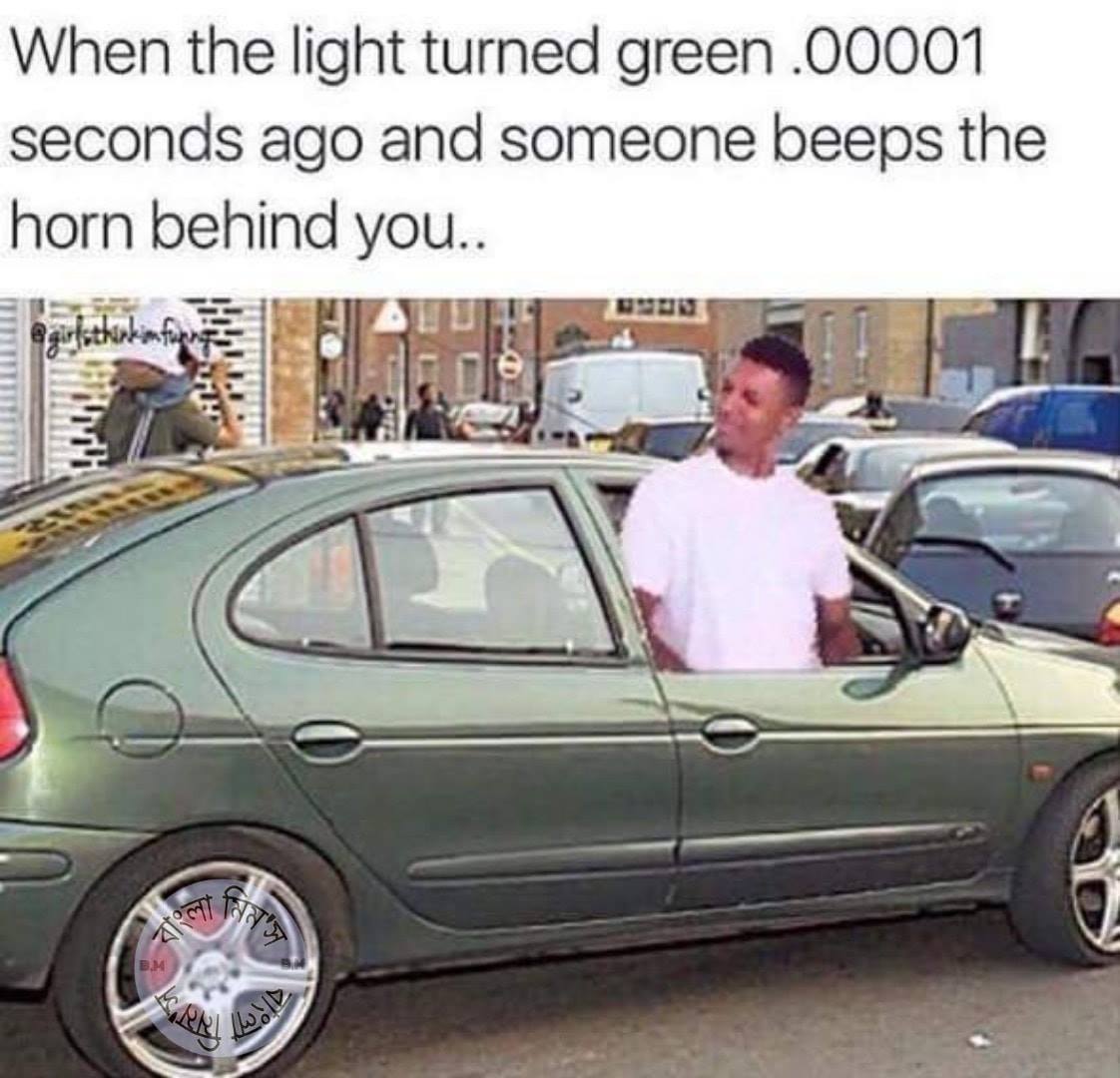}%
\end{subfigure}%
\begin{subfigure}{.5\columnwidth}
\centering
  \includegraphics[height=3cm]{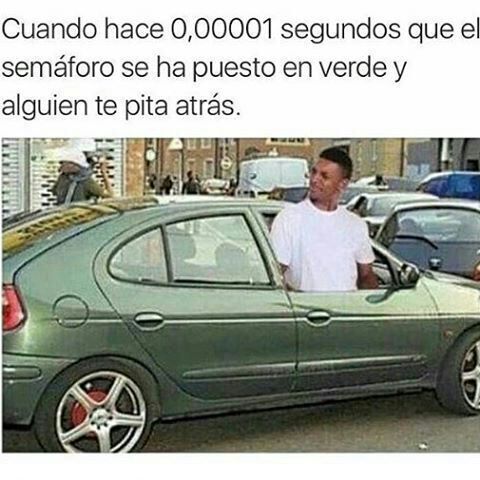}%
\end{subfigure}%
\caption{Example Memetic Image meme \textit{Translation Pair} with textual content in English and Spanish, respectively.}
\label{fig:translation}

\end{figure}

We limited our analysis to meme datasets that are reported to contain only monolingual English memes. However, since our protocol is based on language-agnostic characteristics shared by all multimodal media (e.g., shared visual features, differences in textual content), we posit that our approach can be applied to identify non-English or multilingual memes given annotators who are fluent in the relevant language(s). Additionally, we foresee two linguistic relationships between candidate memes that are absent from a monolingual meme corpus: (1) translation pairs, where two visually identical memes contain semantically identical textual content in different languages, and (2) transliteration pairs, where two visually identical memes contain semantically and linguistically identical textual content in different scripts. Consider the translation pair in Fig. \ref{fig:translation} (the English version was sampled from the \citet{memosen} dataset). One would need to determine whether semantically identical memes in different languages qualify as distinct from each other or not.

An issue that stems from current datasets containing both memetic and non-memetic content is that the role memetic elements play in how memes convey meaning remains unrecognised. Since these elements gain meaning through the memetic process, dedicated knowledge mining and representation approaches are needed to incorporate knowledge of memetic elements into meme classifiers. Recent works that extract and represent large collections of memetic elements \cite{IMKG,sheratt} suggest that knowledge of these elements would likely not be gleaned from a finite dataset of sample memes.

Throughout this paper, we argue that meme classification is a distinct and more complex task than the classification of non-memes. However, we could not demonstrate this difference because the existing meme datasets are not composed exclusively of memes (see Table \ref{tab:results}). As such, we view meme identification as a key step in creating meme classification datasets and advancing meme affective classification approaches. Using our proposed protocol, we plan to develop a dataset that consists only of memes. With such a dataset, we aim to quantitatively assess the difference in performance of state-of-the-art approaches to meme classification versus classification of multimodal content in general.

We also plan to leverage the inherent nature of memes to form genres \cite{memescape} to collect entire groups of visually similar but distinct memes during the \textit{reverse image search} step within our protocol. Within such groups, our aim is to collect memes that serve as contrastive examples of one another: memes that are very similar but contain small differences that allow each to express a different affect. This enables the training of meme classifiers using a supervised contrastive learning approach \cite{contrastive} to capture how the interaction between memetic and novel elements create memes that convey different affects.

 \section{Conclusion}
In this work, we provided a definition for multimodal Internet Memes based on well-established perspectives on memetics from both philological and computational research. We proposed a meme identification model to distinguish memes from non-memetic content. As part of this model, we developed a typology of memes to identify which component of a meme is memetic. We extended this identification model into a protocol that allows meme dataset authors to identify memes in a repeatable and verifiable manner, without relying on personal judgement or familiarity with meme culture. While current meme datasets incorporate some additional effort to separate memes from non-memes, none reported any methodology or criteria used for meme identification. Applying our protocol to 7 meme classification datasets revealed that current meme datasets include both memetic and non-memetic content. Finally, we argued for the creation of meme datasets that exclusively contain memes as a key step towards much needed advancements in meme classification tasks.


\label{sec:reference_examples}
\fontsize{9pt}{10pt} \selectfont
\bibliography{aaai22}


\clearpage

\appendix
 
\section{File Sampling Python Script}
The following are Python functions written to sample candidate memes from each f the dataset analysed in Table \ref{tab:results}.\\[5pt]

\begin{lstlisting}[language=Python,numbers=none]
# Functions to randomly sample candidate memes from folders containing meme datasets.
# Assumes each datasets to comprise either a folder containing image files or a folder of subfolders of image files.
from pathlib import Path
from os import mkdir,listdir
from os.path import join,exists, isdir, basename
import shutil
from typing import List


# Filter non-image files from list of filepaths
def filter_images(files: List[Path]):
    from PIL import Image
    for f in files:
        try:
            Image.open(f).verify()
        except Exception:
            files.remove(f)
    return files

#  Randomly select N number of files from list of filepaths
def random_file_selection(files: List[Path], n: int):
    import random
    return [f for f in random.choices(files,k=n)]

# Randomly select N samples from a dataset
def sample_dataset(input_dir: Path, output_dir: Path, n_samples: int = 200):
    if not exists(output_dir):
        mkdir(output_dir)
    dires = listdir(input_dir)
    # Checks if dataset is split into multiple subfolders.
    if all([isdir(join(input_dir,dire)) for dire in dires]):
        # Collates files from all subfolders
        all_files = []
        for dire in dires:
            for file in listdir(join(input_dir,dire)):
                all_files.extend(join(input_dir,dire,file))
        # Filter for image files only
        img_files = filter_images(all_files)
        # Sample with equal probability ragardless of file source
        samples = random_file_selection(img_files,n_samples)
        for f in samples:
            shutil.move(f, join(output_dir, basename(f)))
    else:
        assert all([not isdir(dire) for dire in dires])
        files = [join(input_dir,f) for f in dires]
        # Filter for image files only
        img_files = filter_images(files)
        # Sample N files
        samples = random_file_selection(img_files,n_samples)
        for f in samples:
            shutil.move(f, join(output_dir, basename(f)))


        
\end{lstlisting}

\section{License Types of Evaluated Datasets}
The following lists the publications and license for each of the dataset used in our Evaluation section.

\begin{table}[ !ht]
    \centering
    \begin{tabular}{|p{0.4\columnwidth} | >{\raggedright\arraybackslash}p{0.5\columnwidth} |  }
    
         \hline
         \textbf{Dataset} & \textbf{License Type}  \\
         \hline
         \citet{hateful_data} &  Proprietary by Facebook, Inc. \\
         \hline
         \citet{fersini} & CC BY-NC-SA 4.0 \\
         \hline
         \citet{harmeme} & BSD \\
         \hline
         \citet{totaldefmemes} & Unspecified but mentions ``non-commercial research purposes''\\
         \hline
         \citet{propmeme} & BSD \\
         \hline
         \citet{memo1_report} & CC BY 4.0\\
         \hline
         \citet{memo2_report} & CC BY 4.0\\
         \bottomrule
    \end{tabular}
    \caption{Distribution license of each Meme Classification Dataset that we sampled for evaluation.}
    
\end{table}

\vfill \eject
\section{Additional Dataset Details}
The following lists the Classification Labels, Size and Meme Sources for each of the Dataset assessed in this work.

\begin{table}[ !ht]
\centering

    \begin{tabular}{|>{\raggedright\arraybackslash}p{0.25\columnwidth} | >{\raggedright\arraybackslash}p{0.4\columnwidth} | p{0.18\columnwidth} | }
    \hline
     \textbf{Dataset}&\textbf{Classification Task} & \textbf{\# Memes}    \normalsize
 \\
 
        \hline

         \citet{Miliani2020DANKMEMESE}&
         \tabitem Meme vs NotMeme\newline \tabitem HateSpeech \newline  \tabitem Event Clustering &
 
         2,361
         \\
         \hline
        
         \citet{memo1_report} &
         \tabitem Sentiment \newline 
         \tabitem Sarcasm\newline 
         \tabitem Humour\newline 
         \tabitem Offensiveness\newline 
         \tabitem Motivational        &
          10,000  
 \\
 \hline
        \citet{memo2_report}&\tabitem Sentiment \newline 
         \tabitem Sarcasm\newline 
         \tabitem Humour\newline 
         \tabitem Offensiveness\newline 
         \tabitem Motivational&
         10,000

         \\
         \hline
         \citet{hateful_data}&  
         Hate speech  &
         10,000

         \\
         \hline
         \citet{harmeme} &
         \tabitem Harmfulness\newline 
         \tabitem Harm Target &
        3,544

         \\
         \hline
         \citet{propmeme}&
         \tabitem Propaganda\newline Technique &
                 2,488 
         \\
         \hline
 
         \citet{tamilmeme}& \tabitem Trolling &
         2,669 
         \\

         \hline             
        
         \citet{totaldefmemes} &\tabitem Defence Topic \newline \tabitem Stance &
         5,301  
         \\
         \hline


          \citet{fersini}&\tabitem Misogyny \newline \tabitem Misogyny type &
         5,500
         \\
         \hline
         \citet{memo3_data}&\tabitem Sentiment \newline 
         \tabitem Sarcasm\newline 
         \tabitem Humour\newline 
         \tabitem Offensiveness\newline 
         \tabitem Motivational&
         10,000
         \\
         \hline
         \citet{memosen}&\tabitem Sentiment &
         4,417 
         \\
         \hline

      \citet{mute} &\tabitem Hate speech &
         4,158
         \\
         \hline
    \end{tabular}
    \caption{Classification tasks and size of the Meme Classification Datasets Evaluated.}
    
\end{table}

\end{document}